\documentclass{ar-1col}

\newcommand{\ORIauthor}{Gammell and Strub}
\newcommand{\ORIshortauthor}{Gammell and Strub}
\newcommand{\ORItitle}{Asymptotically Optimal Sampling-Based Motion Planning Methods} %
\newcommand{\ORIshorttitle}{Asymptotically Optimal Sampling-Based Planning}
\jname{Annu.\ Rev.\ Control Robot.\ Auton.\ Syst.}
\jvol{4}
\jyear{2021}
\doi{10.1146/DOI}
\usepackage{amsmath,amssymb,amsfonts,amscd,amsthm}
\usepackage[printonlyused,nolist]{acronym}
\usepackage{lmodern} %
\usepackage[T1]{fontenc} %
\usepackage{microtype} %
\usepackage{url} %
\usepackage{paralist} %

\usepackage[numbers]{natbib} %

\usepackage{graphicx}
\usepackage{mathtools} %

\usepackage[pdftex,colorlinks]{hyperref}
\usepackage{hypernat} %
\usepackage{bookmark} %
\usepackage[all]{hypcap} %

\hypersetup{%
    pdftitle={\ORIauthor{}: \ORItitle{}},
    pdfauthor={\ORIauthor{}},
    pdfkeywords={},
    pdfsubject={},
    pdfstartview=FitH,%
    breaklinks=true,
    colorlinks=true,
    hidelinks
}
\theoremstyle{plain}
\newcounter{math}
\newtheorem{defn}[math]{Definition}
\renewcommand*{\eqref}[1]{Equation \ref{#1}}
\newcommand*{\SectionTitleFormat}[1]{\texorpdfstring{\expandafter\MakeUppercase\expandafter{#1}}{#1}}
\newcommand*{\KFIJRR}[0]{Karaman and Frazzoli \cite{karaman_ijrr11}}
\newcommand*{\JeaIJRR}[0]{Janson et al.\ \cite{janson_ijrr15}}

\newcommand*{\citeParenthetical}[2]{\cite[#1;][]{#2}}
\newcommand{\citeExample}[1]{\cite[e.g.,][]{#1}} %

\mathchardef\mhyphen="2D %
\newcommand{\bbm}{\begin{bmatrix}}
\newcommand{\ebm}{\end{bmatrix}}

\newcommand*{\norm}[2]{\left\| #1 \right\|_{#2}}

\newcommand*{\suchthat}{\;\; \text{s.t.} \;\;}
\newcommand{\nat}[0]{\mathbb{N}}
\newcommand*{\real}[0]{\mathbb{R}}
\newcommand*{\strictPosReal}[0]{\real_{>0}}
\newcommand*{\set}[1]{\left\lbrace #1\right\rbrace}
\newcommand*{\setst}[2]{\set{#1\;\;\middle|\;\;#2}}
\newcommand{\seqset}[1]{\left(#1\right)}
\newcommand{\seqidx}[4]{\seqset{#1_{#2}}_{#2=#3}^{#4}}
\newcommand*{\closure}[1]{\mathrm{cl}\left(#1\right)}
\newcommand*{\TV}[1]{\mathrm{TV}\left(#1\right)}
\newcommand*{\range}[4]{\left#1 #2, #3\right#4}
\newcommand*{\openrange}[2]{\range{(}{#1}{#2}{)}}
\newcommand*{\closedrange}[2]{\range{[}{#1}{#2}{]}}
\newcommand*{\leftopenrange}[2]{\range{(}{#1}{#2}{]}}
\newcommand*{\rightopenrange}[2]{\range{[}{#1}{#2}{)}}
\newcommand*{\cost}[0]{c}
\newcommand*{\pathCostSymb}[0]{\cost}
\newcommand*{\pathCostDerivSymb}[0]{\mathfrak{c}}
\newcommand*{\kinodynSymb}[0]{f}
\newcommand*{\constraintSymb}[0]{g}
\newcommand*{\countIter}[0]{i}
\newcommand*{\dimensionCntrl}[0]{m}
\newcommand*{\dimension}[0]{n}
\newcommand*{\totalSamples}[0]{q}
\newcommand*{\radius}[0]{r}
\newcommand*{\iterS}[0]{s}
\newcommand*{\pathIter}[0]{t}

\newcommand*{\statex}[0]{\mathbf{x}}
\newcommand*{\statey}[0]{\mathbf{y}}
\newcommand*{\statez}[0]{\mathbf{z}}

\newcommand*{\ball}[2]{B\left(#1, #2\right)}

\newcommand*{\homotopy}[0]{H}
\newcommand*{\probSymb}[0]{P}
\newcommand*{\pathIterLimit}[0]{T}
\newcommand*{\cntrlSet}[0]{U}
\newcommand*{\stateSet}[0]{X}

\newcommand*{\clearance}[0]{\delta}

\newcommand*{\regular}[0]{\xi}
\newcommand*{\pathSeq}[0]{\sigma}
\newcommand*{\cntrlSeq}[0]{\psi}

\newcommand*{\pathSet}[0]{\varSigma}
\newcommand*{\cntrlSeqSet}[0]{\varPsi}

\newcommand*{\prob}[1]{\probSymb\left(#1\right)}

\newcommand*{\namedPathSet}[1]{\pathSet_{#1}}
\newcommand*{\freePathSet}[0]{\namedPathSet{\rm free}}
\newcommand*{\connPathSet}[0]{\namedPathSet{\rm start\mhyphen{}goal}}
\newcommand*{\followPathSet}[0]{\namedPathSet{\rm follow}}
\newcommand*{\feasPathSet}[0]{\namedPathSet{\rm feasible}}
\newcommand*{\clearPathSet}[0]{\namedPathSet{\clearance\mhyphen\mathrm{clear}}}
\newcommand*{\clearPathSeqSet}[1]{\namedPathSet{\clearance_{#1}\mhyphen\mathrm{clear}}}
\newcommand*{\bestPathSet}[0]{\pathSet^{*}}
\newcommand*{\aPath}[0]{\pathSeq'}
\newcommand*{\pathAt}[1]{\pathSeq\left(#1\right)}

\newcommand*{\samplePathSet}[0]{\namedPathSet{\totalSamples}}
\newcommand*{\pathCost}[1]{\pathCostSymb\left(#1\right)}
\newcommand*{\pathCostDeriv}[1]{\pathCostDerivSymb\left(#1\right)}

\newcommand*{\kinodynamic}[1]{\kinodynSymb\left(#1\right)}
\newcommand*{\constraint}[1]{\constraintSymb\left(#1\right)}

\newcommand*{\bestPath}[0]{\pathSeq^{*}}
\newcommand*{\bestPathCost}[0]{\pathCostSymb^{*}}
\newcommand*{\bestRobustPathCost}[0]{\bestPathCost_{\clearance\mhyphen\mathrm{clear}}}
\newcommand*{\pathDeriv}[0]{\dot{\pathSeq}}
\newcommand*{\pathDerivAt}[1]{\pathDeriv\left(#1\right)}
\newcommand*{\cntrlAt}[1]{\cntrlSeq\left(#1\right)}
\newcommand*{\pathRange}[0]{\closedrange{0}{\pathIterLimit}}

\newcommand*{\namedSet}[1]{\stateSet_{#1}}
\newcommand*{\obsSet}[0]{\namedSet{\rm invalid}}
\newcommand*{\freeSet}[0]{\namedSet{\rm free}}
\newcommand*{\goalSet}[0]{\namedSet{\rm goal}}

\newcommand*{\namedState}[1]{\statex_{#1}}
\newcommand*{\xstart}[0]{\namedState{\rm start}}
\newcommand*{\xgoal}[0]{\namedState{\rm goal}} %

\usepackage{eso-pic}
\AddToShipoutPictureFG{%
    \AtTextLowerLeft{%
        \raisebox{-3.5\height}{%
            \parbox[b][2\baselineskip][b]{\textwidth + \extramargin}{%
                Posted with permission from the Annual Review of Control, Robotics, and Autonomous Systems, Volume 4\\
                \textcopyright{}~2021 by Annual Reviews, \url{https://www.annualreviews.org/}.
            }%
        }%
    }%
}

\begin{document}
\makeatletter
\newcommand*{\accite}[2]{%
 \expandafter\ifx\csname AC@#1\endcsname\AC@used
   \acs{#1} \citep{#2}%
 \else
   \acl{#1}\acused{#1} \citep[\acs{#1};][]{#2}%
 \fi
}
\newcommand*{\acpcite}[2]{%
 \expandafter\ifx\csname AC@#1\endcsname\AC@used
   \acsp{#1} \citep{#2}%
 \else
   \aclp{#1}\acused{#1} \citep[\acsp{#1};][]{#2}%
 \fi
}
\makeatother

\begin{acronym}
    \acro{BVP}{boundary-value problem}
    \acro{LQR}{linear-quadratic regulator}
    \acro{LTL}{Linear Temporal Logic}
    \acro{STL}{Signal Temporal Logic}
    \acro{SQP}{sequential quadratic programming}

    \acro{BITstar}[BIT*]{Batch Informed Trees}
        \acrodefplural{BITstar}[BIT*]{Batch Informed Trees}
    \acro{EST}{Expansive Space Tree}
    \acro{FMTstar}[FMT*]{Fast Marching Tree}
    \acro{PRM}{Probabilistic Roadmap}
    \acro{PRMstar}[PRM*]{almost-surely asymptotically optimal \acs{PRM}}
        \acrodefplural{PRMstar}[PRM*]{almost-surely asymptotically optimal \acsp{PRM}}
    \acro{RRT}{Rapidly exploring Random Tree}
    \acro{RRG}{Rapidly exploring Random Graph}
    \acro{RRTstar}[RRT*]{almost-surely asymptotically optimal \acs{RRT}}
        \acrodefplural{RRTstar}[RRT*]{almost-surely asymptotically optimal \acsp{RRT}}
    \acro{sPRM}[s-PRM]{simplified \acs{PRM}}
        \acrodefplural{sPRM}[s-PRM]{simplified \acsp{PRM}}

    \acro{ESP}{Estimation, Search, and Planning}
    \acro{ORI}{Oxford Robotics Institute}
    \acro{EPSRC}{Engineering \& Physical Sciences Research Council}
    \acro{UKRI}{UK Research and Innovation}
\end{acronym} %

\title{\ORItitle{}} %
\author{%
    Jonathan D.\ Gammell and Marlin P.\ Strub%
    \affil{\acf{ESP} Research Group, \acf{ORI}, University of Oxford, Oxford, United Kingdom, OX2~6NN; email: gammell@robots.ox.ac.uk, mstrub@robots.ox.ac.uk}%
}
\markboth{\ORIshortauthor{}}{\ORIshorttitle{}} %
\begin{abstract} %
    Motion planning is a fundamental problem in autonomous robotics that requires finding a path to a specified goal that avoids obstacles and takes into account a robot's limitations and constraints.
    It is often desirable for this path to also optimize a cost function, such as path length.

    Formal path-quality guarantees for continuously valued search spaces are an active area of research interest.
    Recent results have proven that some sampling-based planning methods probabilistically converge toward the optimal solution as computational effort approaches infinity.
    This article summarizes the assumptions behind these popular \emph{asymptotically optimal} techniques and provides an introduction to the significant ongoing research on this topic.
\end{abstract}
\acresetall %
\acused{RRTstar}
\acused{PRMstar}

\maketitle

\section{\SectionTitleFormat{Introduction}}\label{sec:intro}

Planning is an important task in a number of fields, including computer science and robotics.
It consists of finding a sequence of valid states (i.e., a path) between specified positions (i.e., a start and goal) in a search space.
Many problems have multiple \emph{feasible} solutions and applications often seek the feasible path that optimizes a cost function (i.e., the \emph{optimal} solution).
A feasible solution in robot motion planning is a path that avoids hazards in the environment (i.e., obstacles) and can be followed by the robot (e.g., is kinodynamically feasible).
The optimal solution minimizes a user-specified path cost, such as actuator effort or path length.

Optimal planning is difficult because there are often a large number of states to consider and it can be computationally expensive to evaluate them.
Graph-search algorithms, such as A* \cite{hart_tssc68}, can search discrete spaces (e.g., graphs) efficiently with strong formal guarantees.
These techniques are guaranteed to find the optimal solution, if one exists, and otherwise return failure (i.e., they are \emph{complete} and \emph{optimal}).
A* is also guaranteed to expand no more states than any other optimal algorithm given the same information \citeParenthetical{i.e., it is \emph{optimally efficient}}{hart_tssc68}.

Robot motion planning search spaces are instead often continuously valued (i.e., infinite sets) since robots can be arbitrarily repositioned in the physical world.
These spaces can be approximated with discrete representations and then searched with graph-search algorithms but the performance will depend on the chosen resolution.
The resulting discrete solutions will only be resolution complete and resolution optimal relative to the continuous problem.

It can be difficult to select a `correct' \textit{a priori} discrete approximation in many continuously valued problems.
Excessively sparse approximations may preclude finding a (suitable) solution but excessively dense ones can be prohibitively expensive to construct and search.
These difficulties are common in robot motion planning where search spaces are often poorly bounded (e.g., planning outdoors), high dimensional (e.g., planning for manipulation), or otherwise expensive to discretize (e.g., kinodynamic systems).

Sampling-based planning algorithms, such as \accite{PRM}{kavraki_tra96}, \accite{EST}{hsu_ijcga99}, and \accite{RRT}{lavalle_ijrr01}, are designed to avoid \textit{a priori} approximations of the search space.
They instead use sampling to interleave aspects of approximation and search until a solution is found.
This leads to better performance than graph-based methods on many problems but makes formal guarantees probabilistic and dependent on the sampling distribution.
Given an appropriate distribution (e.g., uniform sampling), the probability that many of these techniques find a solution, if one exists, goes to one as the number of samples goes to infinity \citeParenthetical{i.e., they are \emph{probabilistically complete}}{kavraki_tra98,hsu_ijcga99,lavalle_ijrr01}.
Until recently, there were no equivalent formal statements about the quality of these solutions.

\KFIJRR{} present the first formal analysis of probabilistic solution quality in popular sampling-based planning algorithms.
They prove that sampling potential states and statically connecting them to the nearest existing vertex, as in \ac{RRT}, gives a zero probability of finding the optimal solution, even with an infinite number of samples.
They prove that algorithms that consider a higher number of connections, such as \accite{sPRM}{kavraki_tra98}, can have a unity probability of asymptotically converging to the optimal solution, if one exists, as the number of samples goes to infinity (i.e., they are \emph{almost-surely asymptotically optimal}).
They present a series of algorithms specifically designed to consider a sufficient number of connections to achieve almost-sure asymptotic optimality efficiently: \ac{PRMstar}, \ac{RRG} and \ac{RRTstar} \cite{karaman_ijrr11}.

These results have motivated significant recent work on quality guarantees for sampling-based planning algorithms.
These include refining the conditions necessary for popular approaches to converge asymptotically to the optimum and designing novel algorithms that find better initial solutions and/or converge faster.
This survey summarizes results and algorithms from the field to present an introduction to this exciting work.

The remainder of this survey is organized as follows.
Section~\ref{sec:back} introduces the asymptotically optimal planning problem with a focus on providing a common set of definitions and assumptions for the literature.
Section~\ref{sec:lit} presents an introductory survey of work in the field, including important theoretical results and effective algorithms.
Section~\ref{sec:fin} provides a closing summary that includes a discussion of ongoing areas of research interest.
\section{\SectionTitleFormat{Sampling-Based Motion Planning}}\label{sec:back}
\begin{figure}[tbp]%
    \centering%
    \includegraphics[page=1,scale=1]{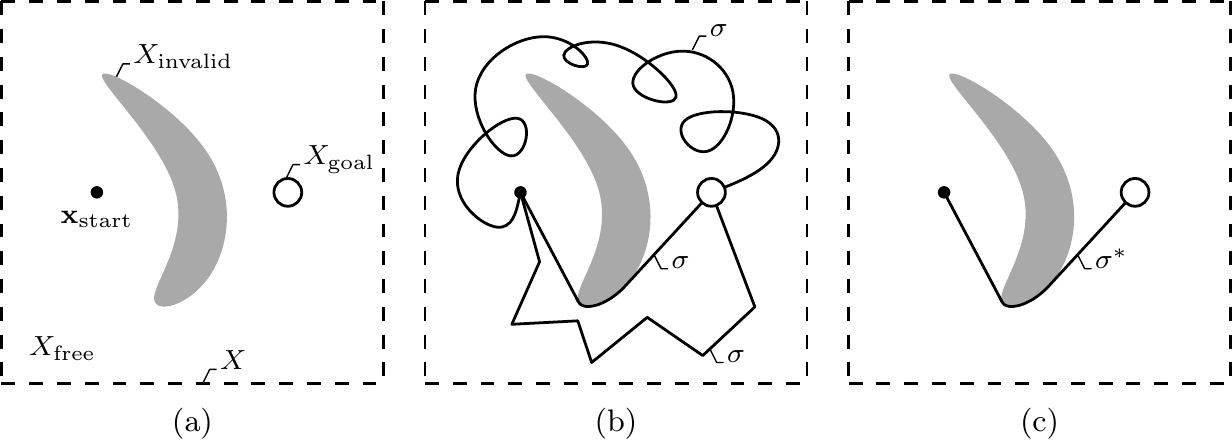}%
    \caption%
    {%
        An illustration of a simple motion planning problem, (a), defined by a search space, $\stateSet$, a start, $\xstart\in\stateSet$, a goal region, $\goalSet\subset\stateSet$, and obstacles, $\obsSet\subset\stateSet$.
        Solutions must pass solely through the set of states not in collision with an obstacle, $\freeSet \coloneqq \stateSet \setminus \obsSet$.
        For this problem there exists an infinite number of feasible solutions, (b), but only one optimal solution with respect to path length, (c).
    }%
    \label{fig:back:prob_defn}%
\end{figure}

The optimal planning problem requires solving the underlying feasible problem (\textbf{Figure~\ref{fig:back:prob_defn}}).
Section~\ref{sec:back:pdefn} presents formal definitions of the planning search space and the feasible and optimal problems.
Section~\ref{sec:back:anal} presents definitions of formal performance guarantees for sampling-based planning algorithms in the form of probabilistic statements on finding solutions to the feasible and optimal motion planning problems.
Section~\ref{sec:back:ass} summarizes common assumptions made to prove these properties for sampling-based planning algorithms.

\subsection{Problem Definitions}\label{sec:back:pdefn}
Planning problems are defined in a spatial representation of the system.
There are a variety of common representations in robot motion planning, including actuator positions (i.e., configuration) and robot pose, possibly including dynamics (i.e., physical state).
This survey refers to all robot representations, without loss of generality, as the search space, $\stateSet$.

\begin{marginnote}%
    \entry{Search space, $\stateSet$}{The space in which the motion planning problem is posed, often a world or configuration space.}%
\end{marginnote}%

A subset of the search space may be invalid for use in a solution, $\obsSet\subset\stateSet$.
This invalid set is not always expressible in closed form and for some problems it can only be defined by a noninvertible function, e.g., $\obsSet \coloneqq \setst{\statex\in\stateSet}{\mathtt{IsInvalid}\left(\statex\right)}$ where $\mathtt{IsInvalid}: \stateSet \to \set{\mathtt{true}, \mathtt{false}}$.
Invalid states in robot motion planning include self collisions, collisions between the robot and the physical world, and other dangerous or undesirable outcomes.
It is often easy to check these conditions for individual states but difficult to enumerate all invalid states, especially when evaluating validity requires mapping between the physical world and configuration space, as in manipulator arms.

The complement of the invalid states is the set of states permitted in a solution, $\freeSet \coloneqq \stateSet \setminus \obsSet$, where $\setminus$ is the set difference.
A planning problem is defined by specifying a start state, $\xstart \in \freeSet$, and a goal state or region, $\xgoal \in \goalSet \subset \freeSet$, in this free space.
In robot motion planning, the start is the initial robot configuration and the goal may be either an individual state (e.g., a mobile robot pose) or set of states (e.g., the set of arm joint angles for a desired end-effector position).

\begin{marginnote}%
    \entry{Free space, $\freeSet$}{The permissible subset of the search space.}%
    \entry{Start state, $\xstart$}{The initial state of a motion planning problem.}%
    \entry{Goal region, $\goalSet$}{The desired final state(s) of a motion planning problem.}%
\end{marginnote}%

A path is a sequence of states through the search space that can be described by a continuous function with bounded variation (i.e., finite length),
\begin{align*}
    \pathSeq : \pathRange \to \stateSet \suchthat \TV{\pathSeq} < \infty,\;\; \forall \pathIter \in \pathRange\; \lim_{\iterS\to \pathIter}\pathAt{\iterS} = \pathAt{\pathIter},
\end{align*}
where $\pathIterLimit\in\strictPosReal$ and $\TV{\cdot}$ is the total variation of the function \cite{karaman_ijrr11}.
The set of paths passing solely through the free space of a problem is the set of free paths,
\begin{align*}
    \freePathSet \coloneqq \setst{\pathSeq\in\pathSet}{\forall \pathIter \in \pathRange\; \pathAt{\pathIter} \in \freeSet},
\end{align*}
where $\pathSet$ is the set of all paths.
The set of paths between the start and goal of a problem is the set of start-goal paths,
\begin{align*}
    \connPathSet \coloneqq \setst{\pathSeq\in\pathSet}{\pathAt{0} = \xstart,\; \pathAt{\pathIterLimit} \in \goalSet}.
\end{align*}
The set of paths executable by the system is the set of followable paths, $\followPathSet$.
This followable set is equivalent to the set of all paths for unconstrained holonomic systems.
Robotic systems can be unconstrained or subject to a variety of constraints and dynamics, such as in kinodynamic systems (Section~\ref{sec:back:ass:kino}).

\begin{marginnote}%
    \entry{Path, $\pathSeq$}{A sequence of states through the search space that can be described by a continuous function with bounded variation.}%
    \entry{Feasible paths, $\feasPathSet$}{The set of all paths connecting the start to goal through free space that can be followed by the robot.}%
\end{marginnote}%

A given problem has a solution if the set of feasible paths,
\begin{align*}
    \feasPathSet \coloneqq \freePathSet \cap \connPathSet \cap \followPathSet,
\end{align*}
is not empty, i.e., $\feasPathSet \not= \emptyset$.
The feasible motion planning problem is then formally defined as the search for a path from the feasible set (Definition~\ref{defn:back:prob:feas}).

\begin{marginnote}%
    \entry{Feasible motion planning}{The search for a feasible path that connects the start and goal of a given problem.}%
\end{marginnote}%
\begin{defn}[Feasible motion planning]\label{defn:back:prob:feas}
    Let $\stateSet \subseteq \real^\dimension$ be the $\dimension$-dimensional search space of the planning problem, $\obsSet \subset \stateSet$ be the set of invalid states, and $\freeSet \coloneqq \stateSet \setminus \obsSet$ be the resulting set of permissible states.
    Let $\xstart \in \freeSet$ be the initial state and $\goalSet \subset \freeSet$ be the set of desired goal states.
    Let $\pathSeq : \pathRange \to \stateSet$ be a continuous function of bounded variation (i.e., a sequence of states) and $\feasPathSet$ be the set of all such paths connecting the start and goal solely through free space that can be executed by the system.

    The feasible motion planning problem is then formally defined as finding any feasible path, $\aPath \in \feasPathSet$, in the problem if a solution exists, i.e., $\feasPathSet \not= \emptyset$, and otherwise returning failure.
\end{defn}

There are often multiple feasible paths to a given problem (\textbf{Figure~\ref{fig:back:prob_defn}b}).
In some applications, any such path is an appropriate solution but other situations require the path that optimizes a specified cost function (\textbf{Figure~\ref{fig:back:prob_defn}c}).
The cost of all paths is assumed to be positive,
\begin{align*}
    \pathCostSymb : \pathSet \to \rightopenrange{0}{\infty} \suchthat \pathCost{\pathSeq} = 0 \iff \forall \pathIter \in \pathRange\; \pathAt{\pathIter} = \statex,
\end{align*}%
such that all feasible solutions to nontrivial problems have finite and strictly positive costs,
\begin{align*}
    \xstart \not\in \goalSet \iff \forall \pathSeq \in \feasPathSet\; \pathCost{\pathSeq} > 0.
\end{align*}
The optimal cost in a specified problem is then given by the infimum of feasible path costs,
\begin{align*}
    \bestPathCost \coloneqq \inf\setst{\pathCost{\pathSeq}}{\pathSeq\in\feasPathSet},
\end{align*}
which defines the set of all optimal paths as
\begin{align*}
    \bestPathSet \coloneqq \setst{\pathSeq\in\feasPathSet}{\pathCost{\pathSeq} = \bestPathCost}.
\end{align*}
Common cost functions in robotics include path length, control effort, and obstacle clearance.

A given problem has an optimal solution if a feasible path exists with optimal cost, i.e., $\bestPathSet \not= \emptyset$.
Problems may not have an optimal solution if the costs of feasible paths are an open set.
The optimal motion planning problem is then formally defined as the search for a path from the optimal set (Definition~\ref{defn:back:prob:opt}).

\begin{marginnote}%
    \entry{Path cost, $\pathCostSymb$}{The cost of a path through the search space. All nontrivial, feasible paths have strictly positive and finite cost.}%
    \entry{Optimal cost, $\bestPathCost$}{The optimal cost of all feasible paths in a problem.}%
    \entry{Optimal paths, $\bestPathSet$}{The set of all feasible paths with optimal cost.}%
\end{marginnote}%

\begin{marginnote}%
    \entry{Optimal motion planning}{The search for a feasible path that connects the start and goal of a given problem such that no other path is better.}%
\end{marginnote}%
\begin{defn}[Optimal motion planning]\label{defn:back:prob:opt}
    Let $\stateSet \subseteq \real^\dimension$ be the $\dimension$-dimensional search space of the planning problem, $\obsSet \subset \stateSet$ be the set of invalid states, and $\freeSet \coloneqq \stateSet \setminus \obsSet$ be the resulting set of permissible states.
    Let $\xstart \in \freeSet$ be the initial state and $\goalSet \subset \freeSet$ be the set of desired goal states.
    Let $\pathSeq : \pathRange \to \stateSet$ be a continuous function of bounded variation (i.e., a sequence of states) and $\feasPathSet$ be the set of all such paths connecting the start and goal solely through free space that can be executed by the system.
    Let $\pathCostSymb : \pathSet \to \rightopenrange{0}{\infty}$ be a cost function such that all nontrivial, feasible paths have finite and strictly positive costs.
    Let $\bestPathSet\coloneqq \setst{\pathSeq \in \feasPathSet}{\pathCost{\pathSeq} = \bestPathCost}$ be the set of all feasible paths with optimal cost, $\bestPathCost$.

    The optimal motion planning problem is then formally defined as finding any feasible path, $\aPath \in \feasPathSet$, in the problem that has optimal cost, i.e., $\aPath \in \bestPathSet$, if an optimal solution exists, i.e., $\bestPathSet \not= \emptyset$, and otherwise returning failure.
\end{defn}

\subsection{Formal Analysis of Sampling-based Motion Planners}\label{sec:back:anal}
Sampling-based motion planners attempt to solve the feasible and optimal motion planning problems by sampling the search space.
These samples are used to approximate and search the problem and can allow the algorithms to be applied to continuously valued spaces without \textit{a priori} finite discretizations.
They also make algorithm performance a function of the number and specific sequence of samples and weaken formal guarantees.

It is common to evaluate sampling-based planning performance as a function of the number of samples probabilistically over all possible realizations of a chosen sampling distribution.
Algorithms with a probability of solving feasible motion planning problems that goes to one with infinite samples are described as probabilistically complete (Definition~\ref{defn:back:anal:pc}) in the sampling-based planning literature \citeExample{kavraki_tra98,hsu_ijcga99,lavalle_ijrr01,karaman_ijrr11}.

\begin{marginnote}%
    \entry{Probabilistically complete}{The probability of finding a solution goes to one as the number of samples approaches infinity.}%
\end{marginnote}%
\begin{defn}[Probabilistic completeness]\label{defn:back:anal:pc}
    A sampling-based motion planning algorithm is said to be \emph{probabilistically complete} if the probability it returns a feasible path, if such a path exists, goes to one as the number of samples goes to infinity,
    \begin{align*}
        \liminf_{\totalSamples\to\infty} \prob{\samplePathSet \not= \emptyset} = 1,
    \end{align*}
    where $\totalSamples$ is the number of samples and $\samplePathSet\subset\feasPathSet$ is the set of feasible paths found by the planner from those samples.
    The probability is calculated over all possible runs of the algorithm, i.e., all realizations of the sampling distribution.
\end{defn}

\KFIJRR{} extend the probabilistic analysis of sampling-based planning algorithms to consider asymptotic quality.
They not only provide a definition of probabilistic convergence to the optimum but also show that \ac{RRT} and variants of \ac{PRM} have zero probability of doing so.
They refer to algorithms that have unity probability of asymptotically converging towards an optimal solution with infinite samples as almost-surely asymptotic optimal (Definition~\ref{defn:back:anal:asao}) and show that variants of \ac{sPRM} and the algorithms \ac{PRMstar}, \ac{RRG}, and \ac{RRTstar} have this property.
This work has since been refined and extended with new analysis (Section~\ref{sec:lit:form}) and to the mathematically weaker concept of being asymptotically optimal in probability (Definition~\ref{defn:back:anal:aop}).
Proving asymptotic optimality almost surely or in probability by definition implies probabilistic completeness.

\begin{marginnote}%
    \entry{Almost-sure asymptotic optimality}{The probability of converging asymptotically to the optimal solution with infinite samples is one.}%
\end{marginnote}%
\begin{defn}[Almost-sure asymptotic optimality]\label{defn:back:anal:asao}
    A sampling-based motion planning algorithm is said to converge asymptotically \emph{almost surely} if it has unity probability of converging asymptotically to an optimal solution, if such a path exists, as the number of samples goes to infinity,
    \begin{align*}
        \prob{\limsup_{\totalSamples\to\infty} \min_{\pathSeq\in\samplePathSet}\set{\pathCost{\pathSeq}} = \bestPathCost} = 1,
    \end{align*}
    where $\totalSamples$ is the number of samples, $\samplePathSet\subset\feasPathSet$ is the set of feasible paths found by the planner from those samples, $\pathCostSymb : \pathSet \to \rightopenrange{0}{\infty}$ is the cost of a path, and $\bestPathCost$ is the optimal solution to the planning problem.
    The probability is calculated over all possible runs of the algorithm, i.e., all realizations of the sampling distribution.
\end{defn}

\begin{marginnote}%
    \entry{Asymptotic optimality in probability}{The probability of a solution being worse than the optimum goes to zero as the number of samples approaches infinity.}%
\end{marginnote}%
\begin{defn}[Asymptotic optimality in probability]\label{defn:back:anal:aop}
    A sampling-based motion planning algorithm is said to converge asymptotically \emph{in probability} if the probability that the best solution cost is more than any positive constant, $\epsilon > 0$, worse than the optimum, if such a path exists, goes to zero as the number of samples goes to infinity,
    \begin{align*}
        \forall \epsilon > 0,\;\; \limsup_{\totalSamples \to \infty} \prob{\left(\min_{\pathSeq\in\samplePathSet}\set{\pathCost{\pathSeq}} - \bestPathCost\right) > \epsilon } = 0,
    \end{align*}
    where $\totalSamples$ is the number of samples, $\samplePathSet\subset\feasPathSet$ is the set of feasible paths found by the planner after those samples, $\pathCostSymb : \pathSet \to \rightopenrange{0}{\infty}$ is the cost of a path, and $\bestPathCost$ is the optimal solution to the planning problem.
    The probability is calculated over all possible runs of the algorithm, i.e., all realizations of the sampling distribution.
\end{defn}

\KFIJRR{} note that all planning algorithms are provably either asymptotically optimal with probability one or zero \citeParenthetical{i.e., almost surely or almost never; Lemma~25}{karaman_ijrr11},\looseness=-1%
\begin{extract}%
    a sampling-based algorithm either converges to the optimal solution in almost all runs, or the convergence does not occur in almost all runs.
\end{extract}%
A corollary of this lemma is that running a sequence of nonasymptotically optimal algorithms, as in Anytime \acsp{RRT} \cite{ferguson_iros06}, is not sufficient to achieve almost-sure asymptotically optimality.
The set of optimal solutions in most practical planning problems has zero measure and therefore all sampling-based planning algorithms have zero probability of sampling an optimal solution in finite time \citeParenthetical{Lemma~28}{karaman_ijrr11},
\begin{extract}%
    no sampling-based planning algorithm can find a solution to the optimality problem in a finite number of iterations.
\end{extract}%

\subsection{Assumptions}\label{sec:back:ass}
Formal analysis of sampling-based planners requires making assumptions about the properties of the optimal motion planning problem.
These commonly include aspects of the search space (Section~\ref{sec:back:ass:space}), solutions (Section~\ref{sec:back:ass:soln}), and cost function (Section~\ref{sec:back:ass:cost}).
These assumptions can also be modified and extended to provide formal analysis of planning performance for kinodynamic systems (Section~\ref{sec:back:ass:kino}).

\subsubsection{Search space assumptions}\label{sec:back:ass:space}
The search space is assumed in \cite{karaman_ijrr11} to be Euclidean and an open unit $\dimension$-dimensional (hyper)cube, $\stateSet \coloneqq \openrange{0}{1}^n$, as in works by Kavraki et al. \cite{kavraki_tra98} and others.
Spaces that are not a unit cube and/or Euclidean must be scaled appropriately and/or behave locally as a Euclidean space.
The free space is then taken as the closed complement of this search space and the invalid set, $\freeSet \coloneqq \closure{\stateSet \setminus \obsSet}$, where $\closure{\cdot}$ denotes the closure of a set.
The closed free set ensures that a feasible path exists with optimal cost for all feasible planning problems, i.e., $\feasPathSet \not= \emptyset \iff \bestPathSet \not= \emptyset$.

\JeaIJRR{} refine these assumptions for planning problems to a goal region to ensure that samples can be drawn near the goal boundary with a nonzero probability.
A goal region is described as $\regular$-regular if its boundary, $\partial\goalSet$, has bounded curvature everywhere.
This is expressed by requiring that every state in the goal boundary also lies on the boundary of a $\regular$-radius subset of the goal region, $\ball{\statey}{\regular} \subseteq \goalSet$, where $\regular>0$, i.e.,
\begin{align}
    \forall \statex \in \partial\goalSet\; \exists \ball{\statey}{\regular}\subseteq\goalSet \suchthat \statex \in \partial\ball{\statey}{\regular},\nonumber
    \shortintertext{where}%
    \ball{\statey \in \stateSet}{\radius \in \strictPosReal} \coloneqq \setst{\statez \in \stateSet}{\norm{\statez - \statey}{2} \leq \radius}\label{eqn:back:ball}
\end{align}
is an $\dimension$-dimensional ball of radius $\radius$ centred at $\statey$ and $\partial$ denotes the boundary of the specified set.

\subsubsection{Solution assumptions}\label{sec:back:ass:soln}
\begin{figure}[tbp]%
    \centering%
    \includegraphics[page=1,scale=1]{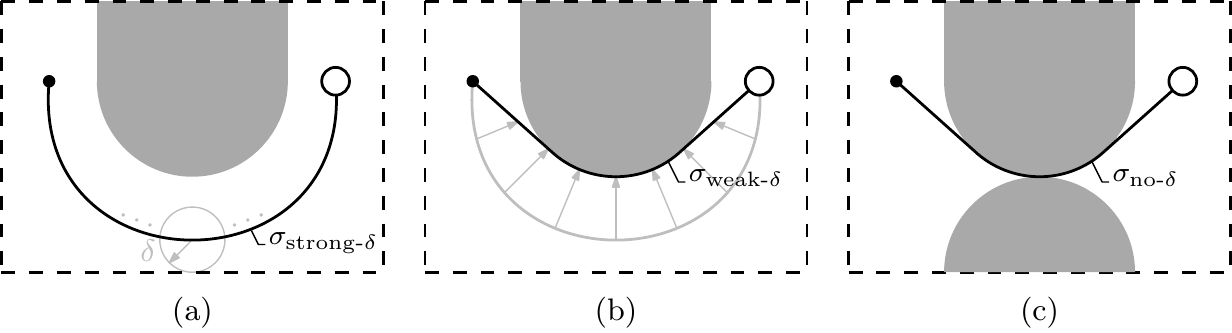}%
    \caption%
    {%
        An illustration of separate paths with strong $\clearance$-clearance (a), weak $\clearance$-clearance (b), and no $\clearance$-clearance (c).
        Probabilistically complete sampling-based motion planners have a probability of finding a feasible path with strong $\clearance$-clearance, e.g., $\pathSeq_{\mathrm{strong}\mhyphen\clearance}$, that goes to one as the number of samples goes to infinity.
        Asymptotically optimal sampling-based motion planners have unity probability of converging to the optimal path with weak $\clearance$-clearance, $\pathSeq_{\mathrm{weak}\mhyphen\clearance}$, as the number of samples goes to infinity.
        Sampling-based motion planners have zero probability of solving problems where all solutions do not have $\clearance$-clearance, $\pathSeq_{\mathrm{no}\mhyphen\clearance}$.
    }%
    \label{fig:back:prob_assume}%
\end{figure}

The probability of sampling a state that could belong to a feasible path is proportional to the measure of the set of all feasible paths relative to that of the sampling domain.
A sufficient condition for this probability to be nonzero when sampling uniformly is for there to exist a feasible path that remains a finite distance from obstacles for its entire length (\textbf{Figure~\ref{fig:back:prob_assume}a}).
Such paths are described as having strong $\clearance$-clearance and the set of all such paths is given by
\begin{align*}
    \clearPathSet \coloneqq \setst{\pathSeq \in \freePathSet}{\exists \clearance>0 \suchthat \forall \pathIter \in \pathRange\; \ball{\pathAt{\pathIter}}{\clearance} \subseteq \freeSet}
\end{align*}
where $\ball{\pathAt{\pathIter}}{\clearance}$ is a $\clearance$-radius ball centred at $\pathAt{\pathIter}$ as defined by \eqref{eqn:back:ball}.
\begin{marginnote}%
    \entry{Strong $\clearance$-clearance path}{A path that is no closer than $\clearance$ from invalid states.}%
    \entry{Robustly feasible problem}{A problem is robustly feasible if it has at least one solution with strong $\clearance$-clearance.}%
\end{marginnote}%

A planning problem containing a feasible path with strong $\clearance$-clearance is described as robustly feasible.
Sampling-based motion planners have zero probability of solving problems where all solutions do not have strong $\clearance$-clearance, i.e., problems that are not robustly feasible (\textbf{Figure~\ref{fig:back:prob_assume}c}).

The set of optimal solutions has zero measure and therefore zero probability of being sampled in most practical problems, even if the problem is robustly feasible.
Asymptotically optimal sampling-based planning algorithms instead converge towards an optimal solution from suboptimal paths (\textbf{Figure~\ref{fig:back:prob_assume}b}).
Optimal solutions often pass infinitely close to obstacles and are described as having weak $\clearance$-clearance if they are homotopic to a strong $\clearance$-clearance path, i.e.,
\begin{align*}
    \exists \homotopy : \closedrange{0}{1} \to \freePathSet \suchthat \homotopy\left(0\right) = \bestPath,\; \homotopy\left(1\right) = \aPath,\; \forall \iterS \in \leftopenrange{0}{1}\; \homotopy\left(\iterS\right) \in \clearPathSet,
\end{align*}
where $\bestPath \in \freePathSet$ is an optimal solution, $\aPath \in \clearPathSet$ is a robustly feasible solution, and $\homotopy$ is a homotopic map between the two.

\JeaIJRR{} note that this homotopy requirement can be ``vacuously satisfied'' (p.~886) and seek to refine the mathematical definition.
They state the assumption as the existence of an infinite sequence of strong $\clearance_\countIter$-clearance paths, $\seqidx{\pathSeq}{\countIter}{1}{\infty}$, such that the limit of this sequence has optimal cost, i.e.,
\begin{align*}
    \exists \seqidx{\pathSeq}{\countIter}{1}{\infty} \suchthat \lim_{\countIter\to\infty} \pathCost{\pathSeq_{\countIter}} = \bestPathCost,\; \forall \countIter \in \nat\; \pathSeq_\countIter \in \clearPathSeqSet{\countIter},
\end{align*}
where the sequence of clearances, $\seqidx{\clearance}{\countIter}{1}{\infty}$, is positive, $\forall\countIter \in \nat\;\; 0 < \clearance_\countIter \leq \clearance$, and converges to zero, $\lim_{\countIter\to\infty}\clearance_\countIter = 0$.

A planning problem with at least one optimal solution that can be continuously transformed to a strong $\clearance$-clearance path is referred to as robustly optimal \cite{karaman_ijrr11} or $\clearance$-robustly feasible \cite{janson_ijrr15}.
Asymptotically optimal techniques cannot converge towards any other optima, i.e., cannot asymptotically solve optimal planning problems that are not robustly optimal or $\clearance$-robustly feasible (\textbf{Figure}~\ref{fig:back:prob_assume}c).

Solovey and colleagues \cite{solovey_ijrr19,solovey_icra20} define the robust optimum, $\bestRobustPathCost$, as the infimum of the $\clearance$-clearance paths,
\begin{align*}
    \bestRobustPathCost \coloneqq \inf\setst{\pathCost{\pathSeq}}{\pathSeq\in\clearPathSet}.
\end{align*}
This cost will be equivalent to the true optimum in problems that contain at least one optimal solution with weak $\clearance$-clearance, i.e., one optimal solution that can be continuously transformed to a strong $\clearance$-clearance path.

\begin{marginnote}%
    \entry{Weak $\clearance$-clearance path}{A path that can be continuously transformed through free space to be no closer than $\clearance$ from invalid states.}%
    \entry{Robustly optimal problem}{A problem is robustly optimal if it has at least one optimal solution with weak $\clearance$-clearance.}%
    \entry{Robust optimum, $\bestRobustPathCost$}{The best cost of all weak $\clearance$-clearance solutions.}%
\end{marginnote}%

\subsubsection{Cost assumptions}\label{sec:back:ass:cost}
Asymptotic convergence towards the optimum requires a well-behaved cost function.
While a variety of assumptions are made about the function in subsequent analysis, the most basic assumption is that it is bounded for paths in free space and monotonic such that the cost of any path cannot be smaller than its subpaths,
\begin{align*}
    \forall \pathSeq_1,\pathSeq_2 \in \pathSet\;\quad \pathCost{\pathSeq_1} \leq \pathCost{\pathSeq_1 \middle| \pathSeq_2}\;\; \text{and}\;\; \pathCost{\pathSeq_2} \leq \pathCost{\pathSeq_1 \middle| \pathSeq_2},
\end{align*}
where $\left.\middle|\right.$ is the concatenation of two paths and is defined as
\begin{align*}
    \forall \pathIter \in \closedrange{0}{\pathIterLimit_{\pathSeq_1} + \pathIterLimit_{\pathSeq_2}}\;\left.\pathSeq_1 \middle| \pathSeq_2\right. \coloneqq
      \begin{cases}
           \pathSeq_1\left(\pathIter\right),         & 0 \leq \pathIter \leq \pathIterLimit_{\pathSeq_1}\\
           \pathSeq_2\left(\pathIter - \pathIterLimit_{\pathSeq_1}\right),     & \pathIterLimit_{\pathSeq_1} < \pathIter \leq \pathIterLimit_{\pathSeq_1} + \pathIterLimit_{\pathSeq_2}
      \end{cases},
\end{align*}
where $\pathIterLimit_{\pathSeq_1}$ and $\pathIterLimit_{\pathSeq_2}$ are the limits of the individual paths.

\begin{marginnote}%
    \entry{Monotonic cost}{A path-cost function where the total cost of a path cannot be smaller than that of any of its constituent subpaths.}%
\end{marginnote}%

\subsubsection{Kinodynamic Systems}\label{sec:back:ass:kino}
The initial formal analysis of asymptotically optimal sampling-based motion planning algorithms focused on geometric motion planning in the absence of dynamics or constraints, i.e., $\followPathSet = \pathSet$.
This analysis has been extended to consider the dynamic systems often found in robotics.
These systems are often described by a dynamical equation that relates the evolution of the state to control inputs (i.e., \textit{kinodynamics}).
This limits the set of followable paths to those that satisfy the differential equation of motion,
\begin{align*}
    \followPathSet \coloneqq \setst{\pathSeq\in\pathSet}{\forall\pathIter\in\pathRange,\;\; \exists\cntrlSeq\in\cntrlSeqSet \suchthat \pathDerivAt{\pathIter}=\kinodynamic{\pathAt{\pathIter},\cntrlAt{\pathIter}}}.
\end{align*}
where $\kinodynSymb\left(\cdot,\cdot\right)$ defines a time-invariant dynamical system and $\cntrlSeqSet\coloneqq\set{\cntrlSeq}$ is the set of all control sequences, $\cntrlSeq : \pathRange \to \cntrlSet$, in the control space of the robot, $\cntrlSet\subseteq\real^\dimensionCntrl$.
The path is fully defined by an initial state, $\pathAt{0} = \xstart$, and the control sequence, which makes kinodynamic motion planning the search for a sequence of controls, $\cntrlSeq'\in\cntrlSeqSet$, that solves the posed problem.

Optimal kinodynamic motion planning problems often seek to minimize the cost of the path and control effort in the form,
\begin{align*}
    \pathCost{\pathSeq,\cntrlSeq} \coloneqq \int_{0}^{\pathIterLimit} \pathCostDeriv{\pathAt{\pathIter},\cntrlAt{\pathIter}}d\pathIter,
\end{align*}
where the integrand, $\pathCostDeriv{\cdot,\cdot}$, is the \emph{cost derivative} and maps from the search and control spaces to a cost,
\begin{align*}
    \pathCostDerivSymb : \stateSet \times \cntrlSet \to \rightopenrange{0}{\infty}.
\end{align*}

Formally analyzing the asymptotic optimality of kinodynamic motion planning algorithms requires additional assumptions that are not summarized here.
These include statements about the controllability of the dynamical system, the nature of the cost function, and other properties of the problem, including whether the dynamical equation can be solved analytically for arbitrary end conditions (i.e., the existence of a `steering' function).
Kinodynamic motion planning is often considered in the presence of kinodynamic constraints (Section~\ref{sec:lit:cnst:kino}).

\section{\SectionTitleFormat{Asymptotically Optimal Sampling-based Motion Planning}}\label{sec:lit}
Asymptotically optimal sampling-based motion planning is a popular research topic.
This section presents an introductory survey of approximately half of the more than 300 academic articles published since 2010, as necessitated by the limits of this article.
While many of these works address multiple questions, general areas of research include refining and extending formal analysis (Section~\ref{sec:lit:form}), improving practical performance (Section~\ref{sec:lit:fast}), supporting constraints and specifications (Section~\ref{sec:lit:cnst}), and applying algorithms to a variety of problems in robotics (Section~\ref{sec:lit:app}).

\subsection{Formal Analysis}\label{sec:lit:form}
A primary area of research interest is refining and extending the formal analysis of asymptotic optimality first presented in \cite{karaman_ijrr11}.
This includes tightening the bounds that guarantee asymptotic optimality almost surely or in probability (Section~\ref{sec:lit:form:bound}), developing relaxed bounds for asymptotic convergence to \emph{near optimal} solutions (Section~\ref{sec:lit:form:near}), and extending analysis to evaluate rates of convergence (Section~\ref{sec:lit:form:rate}).

\subsubsection{Analytic Bounds}\label{sec:lit:form:bound}%
The almost-sure asymptotic optimality presented in \cite{karaman_ijrr11} is a result of considering multiple connections per sample. %
The number of connections necessary are presented as a function of state measure (e.g., volume), state dimension, and the number of existing vertices.
These expressions define either the minimum number of nearest vertices (e.g., $k$-nearest) or the maximum distance to consider (e.g., $r$-disc) when adding new samples.

The computational cost of sampling-based planning algorithms depends on the number of these connections required to be considered for each new sample.
Significant research has worked to develop tighter bounds and/or alternative analysis for other algorithms to reduce the number of connections while maintaining asymptotic convergence to the optimum, sometimes by making more specific assumptions about the planning problem.
Research has included addressing limitations in the original analysis \cite{solovey_icra20}, relaxing asymptotic optimality to convergence in probability \cite{huynh_cdc14,janson_ijrr15}, and developing alternative analysis and refined expressions for connectivity, including with different sampling and graph models \cite{bera_isic13,janson_ijrr18,solovey_ijrr18,solovey_ijrr19}.
It has also investigated necessary conditions for asymptotic convergence when planning for kinodynamic systems \cite{li_ijrr16}, including in an augmented state-cost search space \cite{hauser_tro16,kleinbort_icra20}, and integrated task and motion planning problems \cite{vegabrown_wafr20,shome_wafr20}.
These bounds are often also investigated as part of work focused on other aspects of the planning problem.

\subsubsection{Near Optimality}\label{sec:lit:form:near}%
Asymptotically optimal algorithms converge towards an optimal solution as the number of samples increase but will almost surely not reach it in finite time \citeParenthetical{Lemma~28}{karaman_ijrr11}.
Asymptotically \emph{near-optimal} sampling-based algorithms improve practical performance by instead converging towards a solution that is within a user-specified factor of the optimum.
This relaxed theoretical guarantee reduces computational effort and can result in algorithms that find better solutions faster in finite time and use less computational resources than required to maintain strict asymptotic optimality.

Asymptotic near optimality can be achieved in a variety of ways, including different connectivity expressions \cite{solovey_ijrr18,solovey_ijrr19} and lazily evaluating or removing connections during or after the search \cite{marble_tro13,wang_iros13,dobson_ijrr14,salzman_tro16}.

\subsubsection{Convergence Rate}\label{sec:lit:form:rate}%
Asymptotically optimal sampling-based planning algorithms converge with infinite samples but have no guarantees on their rate of convergence.
Different rates can result in orders-of-magnitude differences in finite-time performance.
Understanding the rates in different situations can help identify useful algorithms for practical problems and also design better planners.

Research includes developing probabilistic bounds on the length of a solution as a function of finite samples for \ac{PRMstar} \cite{dobson_iros13,dobson_icra15} and  \accite{FMTstar}{janson_ijrr15} with both random and deterministic sampling \cite{janson_ijrr18,tsao_icra20}.
It also includes convergence rates for asymptotically near-optimal kinodynamic planners \cite{li_ijrr16} and asymptotically optimal kinodynamic planners built on feasible planning in a state-cost space \cite{hauser_tro16,kleinbort_icra20}.
It has also proven that naive \ac{RRTstar} converges sublinearly (i.e., slower than linearly) in all possible problem or planner configurations when minimizing path length but that focused variants, such as Informed \ac{RRTstar}, can have linear convergence in some situations \cite{gammell_tro18}.

\subsection{Practical Performance}\label{sec:lit:fast}
Another primary area of research is improving the practical ability of asymptotically optimal planning algorithms to find initial solutions quickly and converge rapidly towards the optimum.
This work is important for real robotic systems and may also include formal analysis of the algorithmic improvements.
Research in this area is wide ranging and difficult to classify but includes work on sampling (Section~\ref{sec:lit:fast:samp}), heuristic search (Section~\ref{sec:lit:fast:heur}), lazy computations (Section~\ref{sec:lit:fast:lazy}), hybrid search techniques (Section~\ref{sec:lit:fast:hybrid}), bidirectional search techniques (Section~\ref{sec:lit:fast:bi}), and a variety of other approaches (Section~\ref{sec:lit:fast:misc}).

\subsubsection{Sampling}\label{sec:lit:fast:samp}%
The performance of an individual instance of a sampling-based planning algorithm depends on the sequence of samples used in that specific run.
Sequences may be deterministic \cite{janson_ijrr18,palmieri_ral20,tsao_icra20} or random but expected algorithm performance will depend on the underlying distribution.
Many algorithms use a uniform distribution over the entire search space to ensure that all possible solutions can be found; however, performance can often be improved by increasing the likelihood of sampling (better) solutions.

Samples can be generated in a number of different ways that can reduce search effort, including sampling using motion primitives \cite{settimi_humanoids16}, subspaces or simplified abstractions \cite{kiesel_socs12,brunner_icra13,reid_jfr19,wang_tie20}, potential functions \cite{qureshi_auro16}, and gradient descent \cite{hauer_rss17}.
Performance can also be improved by biasing sampling around approximations of free space \cite{bialkowski_iros13,kim_icra14,meng_robio17,lai_icra19,kang_iros19}, with cross entropy \cite{kobilarov_ijrr12}, machine-learning methods \cite{arslan_iros15,iversen_iros16,lehner_iros17,ichter_icra18,qureshi_iros18,zhang_iros18,kumar_iros19,lai_ral20}, and existing solutions \cite{akgun_iros11,nasir_ijars13}.

Existing solutions to a planning problem limit the search for improvements without loss of generality.
The set of states that could belong to a better solution is the \emph{omniscient set} and sampling it is a necessary condition to improve the solution for many algorithms \citeParenthetical{Lemma~5}{gammell_tro18}.
Knowledge of the omniscient set is equivalent to solving a problem and it is often approximated as an \emph{informed set} whose sampling is also necessary for improvement \citeParenthetical{Lemma~12}{gammell_tro18}.
The utility of informed sets will depend on the precision and accuracy with which they approximate the omniscient set and they are often sampled approximately using heuristics and rejection sampling \cite{akgun_iros11,perez_iros11,otte_tro13,arslan_icra15,kunz_icra16,littlefield_fsr18,yi_icra18,joshi_icra19}.

An informed set applicable to all problems seeking to minimize path length is an $\dimension$-dimensional ellipse defined by the $L^2$ (i.e., Euclidean) norm.
It has been shown that the probability of sampling this set with rejection sampling goes to zero factorially (i.e., faster than exponentially) as state dimension increases \citeParenthetical{Theorem 14}{gammell_tro18}.
This minimum-path-length curse of dimensionality can be avoided without loss of generality by directly sampling the $L^2$ informed set \cite{gammell_tro18}.

\subsubsection{Heuristic Search}\label{sec:lit:fast:heur}%
Graph-search algorithms use estimates of solution cost (i.e., heuristics) to order their search by potential solution quality.
This allows them to prioritize high-quality solutions and avoid unnecessarily low-quality paths which improves both practical and theoretical performance.
Heuristics can also be used to order asymptotically optimal sampling-based planning to find better initial solutions sooner, and converge towards the optimum faster, than uninformed approaches.

Heuristics can be used to order asymptotically optimal sampling-based planning probabilistically \cite{persson_ijrr14} and directly \cite{salzman_icra15a,gammell_ijrr20}.
\accite{BITstar}{gammell_ijrr20} separates approximation from search and uses heuristics to process batches of samples in order of potential solution quality.
This not only searches problems quickly but also allows for extensions from the graph-search literature, including greedy searches \cite{holston_robio17}, heuristic inflation and search truncation to balance exploration and exploitation \cite{strub_icra20a}, and an asymmetric bidirectional search to adaptively estimate and use problem-specific heuristics \cite{strub_icra20b}.
Research into estimating heuristics for motion planning also includes work on kinodynamic systems \cite{paden_ral17}.

\subsubsection{Lazy Search}\label{sec:lit:fast:lazy}%
Sampling-based planning algorithms search problems by sampling the free subspace and connecting samples with edges.
These edges must also pass solely through free space and be followable by the robot (i.e., they must be feasible) for the algorithm to find valid solutions.
Evaluating edge feasibility can be expensive in many problems, especially in the presence of constraints or obstacles that are defined by a noninvertible function of state.

Lazy algorithms reduce computational cost and improve real-time performance by delaying edge evaluations until necessary to find and/or improve a solution.
Edges may be evaluated only when they belong to the best candidate solution \cite{hauser_icra15}, in order of potential solution quality \cite{gammell_ijrr20}, when necessary to satisfy optimality bounds \cite{salzman_tro16}, or to estimate heuristics \cite{salzman_icra15a,strub_icra20b}.
Lazy collision checking can also be adapted as collision information is gathered to reduce false negatives \cite{kim_ur18}.

\subsubsection{Hybrid Search}\label{sec:lit:fast:hybrid}%
Asymptotically optimal sampling-based planning algorithms converge towards the global optimum but have a zero probability of finding it in finite time for most practical problems \citeParenthetical{Lemma~28}{karaman_ijrr11}.
Local search methods, such as path simplification \cite{luna_icra13} or local optimization \cite{zucker_ijrr13}, may find local minima in finite time but provide no global guarantees.
Hybrid search algorithms improve the global convergence of sampling-based planning algorithms by including local optimization in the search.

The balance between global search and local optimization varies across hybrid algorithms.
Some techniques apply optimizers to improve connections between samples \cite{choudhury_icra16,suh_tro17,kim_auro20} and/or minimize solutions \cite{nasir_ijars13,kim_isr18,kuntz_isrr20} during global search.
Others are designed to use sampling-based exploration to explore homotopy classes as initial conditions for optimization methods \cite{kim_iros19}.

\subsubsection{Bidirectional Search}\label{sec:lit:fast:bi}%
Probabilistically complete bidirectional sampling-based planning algorithms, such as \acs{RRT}-Connect \cite{kuffner_icra00}, are effective techniques for feasible planning problems.
Bidirectional variants of asymptotically optimal algorithms apply similar approaches to the optimal planning problem.
This often finds initial solutions faster but incorporating bidirectional search into asymptotic convergence to the optimum can be more complex.

Bidirectional asymptotically optimal planning includes direct implementations of ``\acs{RRTstar}-Connect'' \cite{akgun_iros11,klemm_robio15} and modifications to improve asymptotic convergence by limiting the `Connect' heuristic \cite{jordan_tech13,qureshi_ras15} and using informed sampling \cite{burget_iros16}.
They have also been used for manifold constraints \cite{jaillet_ras13}, replanning in the presence of dynamic obstacles \cite{boardman_allerton14}, and to estimate heuristics adaptively during search \cite{strub_icra20b}.
A bidirectional \ac{FMTstar} also includes research on stopping conditions for bidirectional marching methods \cite{starek_iros15}.

\subsubsection{Other Search Improvements}\label{sec:lit:fast:misc}%
Not all methods to improve the practical performance of asymptotically optimal sampling-based planning algorithms fit neatly into the previous descriptions.
A variety of other techniques to improve asymptotically optimal planning include switching between nonasymptotically optimal and asymptotically optimal algorithms \cite{alterovitz_icra11,shan_iv14} and more thoroughly exploiting sampled information \cite{arslan_icra13,arslan_icra15}.
It also includes using computational resources more effectively, such as techniques designed for parallel computing \cite{otte_tro13,arslan_cdc16}.
\looseness=-1%

\subsection{Constrained Planning Problems}\label{sec:lit:cnst}
Robot motion planning problems often require solutions that not only avoid obstacles but also satisfy other platform and/or task constraints.
These can include manifold (Section~\ref{sec:lit:cnst:man}), nonholonomic (Section~\ref{sec:lit:cnst:noholo}), and kinodynamic (Section~\ref{sec:lit:cnst:kino}) constraints and high-level task specifications (Section~\ref{sec:lit:cnst:form}).

\subsubsection{Manifold Constraints}\label{sec:lit:cnst:man}%
Many robot motion planning problems require solutions to both avoid obstacles and satisfy geometric constraints, such as maintaining the orientation of a manipulator end effector.
These kinematic or holonomic constraints are a function of state, e.g., $\constraint{\statex} = 0$, and limit solutions to a lower dimensional manifold of the original problem.
This manifold has zero measure and therefore zero probability of being sampled from the full search space.
\begin{marginnote}%
    \entry{Manifold constraint}{A constraint on the system state which restricts solutions to a lower dimensional manifold in the original search space.}%
\end{marginnote}%

Problems that cannot be reparameterized onto the constraint manifold require planning techniques that can map their search to the implicit manifold.
Research on asymptotically optimal algorithms to do so includes using continuation techniques \cite{jaillet_ras13} and decomposition into finite subspaces \cite{vegabrown_wafr20}.
It also includes projection-{} and continuation-based methods that allow many types of general asymptotically optimal motion planners to be used directly on implicit manifold configuration spaces \cite{kingston_ijrr19}.

\subsubsection{Nonholonomic Constraints}\label{sec:lit:cnst:noholo}%
Many real-world robots have restrictions on their motion, such as skid-steer and Ackermann-steer wheeled robots that cannot move laterally.
These nonholonomic constraints are an inseparable function of the state and its derivatives, e.g., $\constraint{\statex, \frac{d\statex}{dt}, \frac{d^2\statex}{dt^2}, \ldots} = 0$, and limit the connectivity of the search space.
These problems can often be solved by treating the nonholonomic constraints as additional obstacles but better planning performance can be achieved by actively incorporating the constraints into the search.
Many of these systems must also consider kinodynamic constraints (Section~\ref{sec:lit:cnst:kino}).
\begin{marginnote}%
    \entry{Nonholonomic constraint}{A constraint coupling the system state and its derivatives which reduces the connectivity of the search space.}%
\end{marginnote}%

Research has developed distance functions for a variety of nonholonomic vehicles \cite{karaman_icra13,park_iros15} and a general framework to assess asymptotically optimal planning for driftless systems \cite{schmerling_icra15}.
It also includes work on asymptotically optimal feedback planning \cite{yershov_ijrr16}, planning in vector flow fields \cite{palmieri_icra17}, and deterministic sampling for driftless systems \cite{palmieri_ral20}.

\subsubsection{Kinodynamic Constraints}\label{sec:lit:cnst:kino}%
The motion of robots in the real world is governed by differential equations relating control inputs (e.g., forces) to accelerations (Section~\ref{sec:back:ass:kino}).
These kinodynamics change the connectivity of the search space and any kinodynamic constraints that limit control inputs or rates of change, e.g., $\frac{d^2\statex}{dt^2}<a_{\mathrm{max}}$, further reduce the set of feasible paths.
This kinodynamic planning is further complicated for systems where the differential equations of motion cannot be solved in closed form for arbitrary end conditions.
The absence of analytical solutions to these two-point \acp{BVP} require additional considerations during asymptotically optimal planning, such as numerical approximations or shooting methods.
Many of these systems must also consider nonholonomic constraints (Section~\ref{sec:lit:cnst:noholo}).
\begin{marginnote}%
    \entry{Kinodynamic constraint}{A limit on the state derivatives or control inputs of a kinodynamic system which reduces the connectivity of the search space.}%
\end{marginnote}%

This popular area of research includes analyzing kinodynamic asymptotic optimality \cite{karaman_cdc10,schmerling_icra15,schmerling_cdc15} and extending \ac{RRTstar} to kinodynamic systems \cite{jeon_cdc11,perez_icra12,karaman_icra13,goretkin_icra13,webb_icra13,ha_cdc13,arslan_icra17,sakcak_auro19}.
A wide variety of kinodynamic planning techniques exist, including solving two-point \acp{BVP} with \acpcite{LQR}{perez_icra12,goretkin_icra13}, fixed-final-state free-final-time controllers \cite{webb_icra13}, successive approximation \cite{ha_cdc13}, closed-loop prediction \cite{arslan_icra17}, and precomputed motion primitives \cite{sakcak_auro19}.
Other asymptotically optimal kinodynamic planning algorithms use \ac{SQP} with \ac{BITstar} \cite{xie_icra15} and analytic steering solutions \cite{jeon_icra15} and numerical approximations \cite{yershov_ijrr16} with marching methods.
Kinodynamic techniques have also been accelerated with heuristics and informed sampling \cite{kunz_icra16,littlefield_fsr18,littlefield_iros18,yi_icra18}.

Solving or approximating two-point \acp{BVP} is impractical or impossible in some problems.
Research on these applications has extended \acs{RRT}-style shooting methods to asymptotically optimal kinodynamic planning \cite{abbasi-yadkori_iros10,li_ijrr16}, including with an augmented state-cost search space that does not require rewiring \cite{hauser_tro16,kleinbort_icra20}.
Two-point \acp{BVP} can also be avoided with generalized label correcting methods \cite{paden_wafr20}.

\subsubsection{High-level Task Specifications}\label{sec:lit:cnst:form}%
Many motion planning problems require solutions to both avoid obstacles and satisfy a number of high-level or task-specific constraints.
These may include the rules of the road for self-driving cars, a time-dependent sequence of tasks for a warehouse robot, or any number of other complex temporal relationships.
These requirements are often defined in high-level specification languages and must be evaluated in parallel to searching for a collision-free path.

A number of different specification languages have been used with asymptotically optimal sampling-based planners, including $\mu$-calculus \cite{karaman_acc12}, process algebra \cite{varricchio_icra14}, \accite{LTL}{cho_ral17,oh_cdc17,zhang_ast20}, finite \ac{LTL} \cite{reyescastro_cdc13}, and \accite{STL}{vasile_iros17}.
This research allows for algorithms to find paths that satisfy high-level constraints and, when no satisfying solution exists, find paths that reach the goal while violating a minimum number of specifications \cite{reyescastro_cdc13}.

\subsection{Applications and Further Extensions}\label{sec:lit:app}
There are a number of interesting and challenging extensions of the motion planning problem in robotics.
These include complex cost functions (Section~\ref{sec:lit:app:cost}) and planning for multirobot (Section~\ref{sec:lit:app:mrobot}) and multimodal (Section~\ref{sec:lit:app:mmode}) systems.
It also includes considering state and/or measurement uncertainty during planning (Section~\ref{sec:lit:app:uncert}) and replanning when the environment changes or new information is available (Section~\ref{sec:lit:app:replan}).

Asymptotically optimal planning algorithms are also used in a variety of interesting applications.
These include finding optimal solutions to the filtering \cite{chaudhari_cdc12} and stochastic control problems \cite{huynh_ijrr16,huynh_cdc12,chaudhari_acc13,huynh_cdc14} and planning for a number of specific situations, including pursuit evasion games \cite{karaman_wafr10}, simultaneous planning and execution \cite{karaman_icra11}, moving goals \cite{liu_icma18}, operating in vector flow fields \cite{palmieri_icra17,to_icra20}, autonomous driving at high speed \cite{jeon_acc13}, autonomous driving for passenger comfort \cite{shin_iros18}, cable-suspended parallel robots \cite{xiang_jmr20}, to reduce radioactive exposure \cite{chao_net19}, and very many more.

\subsubsection{Different Cost Functions and Objectives}\label{sec:lit:app:cost}%
Many planning problems seek to optimize a more complex cost than path length, such as maximizing minimum clearance to obstacles, and objectives defined by cost maps or combinations of cost functions.
These objective functions may not meet the assumptions used to first prove asymptotically optimality (Section~\ref{sec:back:ass}) or the resulting refinements (Section~\ref{sec:lit:form:bound}).

Asymptotically optimal algorithms for these situations include planning on problems defined by cost maps \cite{devaurs_tase16}, finding Pareto optimal solutions to multiobjective problems \cite{yi_ijcai15}, and minimizing bottleneck (i.e., maximum state) cost \cite{solovey_ijrr19}.
Specific applications include asymptotically optimal inspection in terms of path length and coverage \cite{fu_rss19} and planning for groups of vehicles to optimize connectivity, surveillance, and path length \cite{rahman_case16}.

A reoccurring complex planning objective in robotics is information maximization.
Informative path planning problems require systems to measure their environment while navigating to any specified goals.
Asymptotically optimal algorithms have been developed and demonstrated for this problem in environmental monitoring \cite{hollinger_ijrr14,jadidi_ijrr19} and mapping \cite{sayremccord_iros18}.
\looseness=-1%

\subsubsection{Multirobot Systems}\label{sec:lit:app:mrobot}%
The complexity of path planning is directly related to the dimensionality of the search space.
This dimensionality increases quickly in systems consisting of multiple robots.
These problems can often be solved naively by treating each robot independently but this may preclude solutions to some problems that require coordination.

Research includes identifying conditions for multirobot asymptotic convergence and developing techniques to solve coupled multirobot problems with the tensor product of their individual spaces \cite{shome_auro20}.
Applications also include groups of vehicles optimizing complex cost functions \cite{rahman_case16}, two-arm manipulation \cite{shome_humanoids17}, and cooperative aerial transport \cite{lee_tase18}.

\subsubsection{Multimodal Systems}\label{sec:lit:app:mmode}%
Some complex robotic systems have multiple configurations that are best represented as discrete modes (e.g., different legged-robot gaits).
Planning problems for these multimodal systems have search spaces with both continuously valued and discrete dimensions.
These problems can often be solved with general sampling-based planners but techniques designed to exploit the mixed nature of the search space can be more efficient.

Work to develop better asymptotically optimal planning algorithms for these situations includes extending roadmap planners to the multimodal configuration space of robot and object manipulation \cite{schmitt_icra17}.
It also includes investigating the conditions for asymptotically optimal convergence in multimodal integrated task and motion planning problems \cite{shome_wafr20} and the application of multirobot techniques to this problem \cite{shome_mrs19}.

\subsubsection{Uncertainty}\label{sec:lit:app:uncert}%
Real-world robotic systems use noisy sensors to measure their environment.
This creates uncertainty about their position relative to both obstacles and the goal and makes the feasible and optimal motion planning problems probabilistic.
This \emph{belief space planning} seeks a solution that safely reaches the goal with high probability given the system uncertainty.

Asymptotically optimal algorithms in belief space find the minimum cost path for a noisy system and/or in an uncertain environment that has a bounded probability of collision \citeParenthetical{i.e., chance constraint}{bry_icra11,luders_gnc13,bopardikar_ijrr16,sun_isrr16,virani_acc16,yang_iros16} or is guaranteed to be safe \cite{luders_acc14}.
Research has also studied the conditions required for optimality in belief space and shown that this property cannot be achieved for several cost functions \cite{shan_iros17} and investigated better distance functions to improve planning \cite{littlefield_isrr18}.

\subsubsection{Unknown and/or Dynamic Environments}\label{sec:lit:app:replan}%
Real-world robotic systems often operate in environments where the known presence and position of obstacles can change over time as a result of moving objects or sensor range limits.
These changes may invalidate previously feasible solutions and require the system to solve an updated version of the original planning problem.
This problem can be solved more efficiently with planning algorithms that reuse the previous search effort.

Asymptotically optimal techniques to replan efficiently when new information becomes available include using a bidirectional search to facilitate updates \cite{boardman_allerton14} and continuously refining and repairing a search during execution \cite{otte_ijrr16,yang_iros16,chen_auro19}.
Work has also investigated achievable notions of optimality specific to incrementally revealed environments that result in policies guaranteed to be collision free \cite{janson_rss18}.

\section{\SectionTitleFormat{Conclusion}}\label{sec:fin}

Sampling-based motion planning algorithms are powerful tools for searching continuously valued spaces, as often found in robotics.
They use samples to approximate and search the space and many popular algorithms have a unity probability of finding a solution, if one exists, with an infinite number of samples (i.e., they are probabilistically complete; Definition~\ref{defn:back:anal:pc}).
The quality of the solutions returned by these algorithms remained an open question until recently.

\KFIJRR{} analyze the quality of solutions found by popular sampling-based algorithms and prove that most have zero probability of finding an optimal solution, even with infinite samples.
They provide efficient versions of these popular algorithms that instead converge asymptotically to the optimal solution with infinite samples almost surely over all realizations of a sampling distribution (i.e., they are almost-surely asymptotically optimal; Definition~\ref{defn:back:anal:asao}).

Most asymptotically optimal algorithms converge towards the optimal solution incrementally with additional samples.
This anytime performance avoids the difficulties of approximating a continuously valued search space \textit{a priori} to its search.
The sampling instead interleaves approximation and search and the algorithms can be run for a given computational budget or until a suitable solution is found.

Extending asymptotically optimal planning has become an important and increasingly popular area of theoretical and practical research.
Theoretical work has refined the necessary conditions for both asymptotic optimality almost surely and in probability (Definition~\ref{defn:back:anal:aop}), investigated asymptotic near-optimality, and analyzed rates of convergence.
Practical work has developed a wide-variety of techniques to find better initial solutions sooner and converge towards the optimum faster.

Asymptotically optimal motion planning algorithms have been adopted and used on a number of robotic systems and problems.
These include mobile ground robots, multirotor and fixed wing aerial vehicles, manipulator systems, and many others.
These systems may operate independently or as part of a coordinated group and they may be unconstrained or have manifold, nonholonomic, kinodynamic, or high-level task constraints.
The planning algorithms may need to optimize complex cost functions and account for unknown environments, moving obstacles, and measurement uncertainty in their plans.

Ongoing research will likely continue to expand the applicability and performance of these popular motion planning algorithms.
This will include additional theoretical results, including perhaps further refinement to the necessary conditions for asymptotic optimality and investigations on the convergence rate of the resulting algorithms.
It will also continue to include wide ranging efforts to improve practical search performance and applicability to real-world robotic problems.

\begin{summary}[\SectionTitleFormat{Summary Points}]
    \begin{enumerate}
        \item Asymptotically optimal sampling-based planning algorithms are applicable to many optimal motion planning problems with continuously valued search spaces.%
        \item These algorithms asymptotically converge to the optimal solution as the number of samples goes to infinity almost surely or in probability over all realizations of an appropriate sampling distribution.%
        \item Many of these algorithms converge in an anytime manner that avoids the difficulty of selecting the correct approximation \textit{a priori} to the search.%
        \item These algorithms have been successfully applied to a number of important problems in robotics, including nonholonomic and kinodynamic systems, in unknown and dynamic environments, and in the presence of execution and measurement uncertainty.%
    \end{enumerate}
\end{summary}

\begin{issues}[\SectionTitleFormat{Future Issues}]
    \begin{enumerate}
        \item Research continues to refine the conditions necessary for asymptotic optimality for a wide range of problems and cost functions.%
        \item There is a large amount of interest in improving practical search performance by finding and improving solutions quickly.%
        \item Potential real-world applications continue to grow and include new challenging problems and environments.%
    \end{enumerate}
\end{issues}

\section*{\SectionTitleFormat{Disclosure Statement}}
The authors are not aware of any affiliations, memberships, funding, or financial holdings that might be perceived as affecting the objectivity of this review.

\section*{\SectionTitleFormat{Acknowledgments}}
Work on this review was supported in part by the University of Oxford and \acl{UKRI} and \acs{EPSRC} through the ``Robotics and Artificial Intelligence for Nuclear (RAIN)'' research hub [EP/R026084/1] and the ``ACE-OPS: From Autonomy to Cognitive assistance in Emergency OPerationS'' international centre-to-centre research collaboration [EP/S030832/1].

\phantomsection%
\addcontentsline{toc}{section}{\SectionTitleFormat{Literature Cited}}%
\bibliographystyle{ar-style3.bst}
\bibliography{TR-2020-JDG001}
\end{document}